\documentclass[10pt,twocolumn,letterpaper]{article}

\usepackage{cvpr}              
\definecolor{cvprblue}{rgb}{0.21,0.49,0.74}
\usepackage[pagebackref,breaklinks,colorlinks,allcolors=cvprblue]{hyperref}
\usepackage{color}

\def\paperID{} 
\def\confName{CVPR}
\def\confYear{2026}

\title{PhySe-RPO: Physics and Semantics Guided Relative Policy Optimization \\ for Diffusion-Based Surgical Smoke Removal}

\author{
Zining Fang$^{1\dagger}$,
Cheng Xue$^{1\dagger *}$,
Chunhui Liu$^{2}$,
Bin Xu$^{2}$,
Ming Chen$^{2}$, and 
Xiaowei Hu$^{3}$\\
$^{1}$School of Computer Science and Engineering, Southeast University, $^{2}$Zhongda Hospital,Southeast University\\
$^{3}$School of Future Technology, South China University of Technology \\
{\tt\small \{znfang, cxue\}@seu.edu.cn}
}

\begin{document}
\maketitle
\begingroup
\renewcommand\thefootnote{$\dagger$}
\footnotetext{Equal contribution.}
\renewcommand\thefootnote{$*$}
\footnotetext{Corresponding author.}
\endgroup
\begin{abstract}
Surgical smoke severely degrades intraoperative video quality, obscuring anatomical structures and limiting surgical perception. Existing learning-based desmoking approaches rely on scarce paired supervision and deterministic restoration pipelines, making it difficult to perform exploration or reinforcement-driven refinement under real surgical conditions. We propose \textbf{PhySe-RPO}, a diffusion restoration framework optimized through \textbf{Physics- and Semantics-Guided Relative Policy Optimization}. The core idea is to transform deterministic restoration into a \textit{stochastic policy}, enabling trajectory-level exploration and critic-free updates via group-relative optimization. A physics-guided reward imposes illumination and color consistency, while a visual-concept semantic reward learned from CLIP-based surgical concepts promotes smoke-free and anatomically coherent restoration. Together with a reference-free perceptual constraint, PhySe-RPO produces results that are physically consistent, semantically faithful, and clinically interpretable across synthetic and real robotic surgical datasets, providing a principled route to robust diffusion-based restoration under limited paired supervision.
\end{abstract}

\section{Introduction}
\label{sec:intro}

Robot-assisted minimally invasive surgery has transformed clinical practice by enabling precise, dexterous manipulation through endoscopic visualization, providing real-time views of the operative field and allowing complex procedures to be performed through small incisions with enhanced control and stability.
However, videos captured during operation are frequently degraded by dense smoke, blur, and non-uniform illumination arising from energy based tissue dissection and light scattering on moist organ surfaces. 
Such degradations obscure fine anatomical details, hinder visual perception, and increase the cognitive load on surgeons, potentially compromising intraoperative safety and procedural outcomes.
\begin{figure}[t] 
    \centering 
    \includegraphics[width=\linewidth,keepaspectratio]{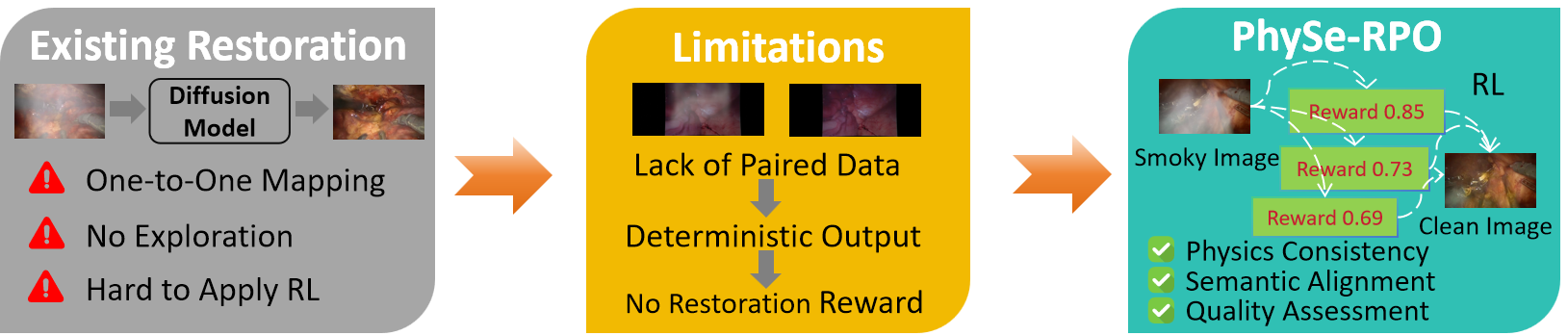}
    \caption{Limitations of existing restoration approaches: lack of paired data, restoration produce a deterministic output making reward learning difficult, and lack of desmoke restoration-oriented rewards. Our PhySe-RPO addresses these issues by turning restoration into a stochastic policy optimization problem and using physics- and semantics-guided rewards to learn effectively from unlabeled real surgical videos. } 
    \label{fig:1}
    \vspace{-20pt}  
\end{figure}

Traditional image desmoking methods, such as Dark Channel Prior (DCP)~\cite{he2010single} and other physics-based approaches derived from the atmospheric scattering model, have shown strong performance in natural image restoration. 
However, these models are less effective in robotic surgical scenes, where lighting is non-uniform, reflections are intense, and textures vary across tissue types. 
As a result, conventional priors often oversmooth structural boundaries or distort color balance when applied to surgical videos. 
Recent advances in deep learning have made it possible to learn complex restoration mappings directly from data, and diffusion-based generative models~\cite{chen2024lightdiff,lu2024mixdehazenet,luo2023controlling,liao2024denoising,fu2024temporal,chang2024lsd3k} have demonstrated impressive results on desmoking tasks. Meanwhile, emerging reinforcement-learning–based frameworks~\cite{shao2024deepseekmath} have shown that paired data is not strictly necessary, as models can be optimized directly from reward signals derived from unpaired images.
Yet, their direct deployment in surgical image restoration remains limited by three key factors: 
(1) the scarcity of large-scale paired smoky to clean surgical datasets for supervised training; and 
(2) Diffusion-based restoration follows a one-to-one mapping with little output diversity, which limits exploration and makes reward-driven refinement challenging under unpaired real surgical conditions.
(3) Current rewards for vision tasks are primarily for generation instead of restoration, as shown in Figure~\ref{fig:1}.

To address these limitations, we reformulate diffusion-based restoration as a stochastic policy optimization problem guided by physics and semantic constraints. 
Specifically, we propose \textit{PhySe-RPO} (Physics- and Semantics-Guided Relative Policy Optimization), a diffusion-based framework that bridges deterministic restoration with reinforcement-guided policy learning. 
Unlike conventional diffusion models that generate a single fixed restoration trajectory for each input, PhySe-RPO introduces controlled stochasticity to enable exploration in the solution space and leverages a reward-driven mechanism to refine the policy toward physically consistent and clinically interpretable restorations. 
The framework integrates three key components: a group relative diffusion policy optimization strategy for critic free reinforcement learning, physics guided rewards via color priors to enforce illumination and chromatic consistency, and visual concept semantic rewards to ensure high level alignment with the clear surgical scene concept. 
Additionally, a reference free quality constraint based on learned image quality surgical datasets.
The main contributions  are summarized as follows:
\begin{itemize}
    \item We present PhySe-RPO, a diffusion-based framework that turns deterministic restoration into stochastic policy optimization, enabling physics- and semantics-guided reinforcement refinement for surgical smoke removal.

    \item We develop a group-relative diffusion policy optimization scheme that converts diffusion restoration into an RL-optimizable stochastic policy, enabling stable critic-free updates and improved robustness on unpaired real data.

    \item We introduce physics-based color priors and a visual-concept semantic reward to jointly enforce illumination fidelity, perceptual realism, and clinically interpretable and reliable reconstruction quality.

    \item We build a robotic surgical dataset combining simulated smoky–clean pairs with real surgical videos, providing a benchmark for surgical scene restoration.
\end{itemize}

\section{Related Work}
\noindent\textbf{Image Desmoking Methods.}

Early image desmoking methods rely on the atmospheric scattering model, where clear scene radiance is recovered by estimating transmission and illumination. Classical priors such as the Dark Channel Prior (DCP)~\cite{he2010single} use handcrafted statistics to invert this model and have shown strong performance on natural images. However, when applied to robotic surgical videos, these priors often break down due to non-uniform endoscopic illumination, strong specular reflections, and heterogeneous tissue textures, leading to oversmoothing and color distortions.

To overcome the limitations of assumptions relying only on physics, learning-based desmoking approaches have been widely explored. GAN-based medical desmoking methods~\cite{wang2023surgical,zhou2022synchronizing,salazar2020desmoking,pan2022desmoke,chen2019smokegcn,hu2021cycle,venkatesh2020unsupervised} learn smoky to clear mappings to restore anatomical visibility. Although effective, GANs still require substantial paired supervision and may hallucinate structures when training data is limited or unpaired. Diffusion models~\cite{Ho2020DDPM,song2020score} have emerged as powerful generative priors for vision tasks at the low level due to their iterative denoising and strong distribution modeling capacity. Diffusion-based desmoking and dehazing methods~\cite{chen2024lightdiff,lu2024mixdehazenet,luo2023controlling,liao2024denoising,fu2024temporal,chang2024lsd3k} achieve impressive clarity and structural fidelity. However, most diffusion pipelines operate under supervised or paired settings, and paired smoky and clean surgical images are scarce. 

\noindent\textbf{Reinforcement Learning in Vision Tasks.}
Reinforcement learning has recently gained traction in vision tasks where reward-driven optimization can replace the need for paired supervision. 
RL has been used for sample selection, segmentation prompt generation~\cite{you2025seg}, and scene understanding in VLM-based agents~\cite{guo2025improving}. 
RL has also begun to influence diffusion models, with recent works~\cite{wang2025simplear,lin2025reasoning,chen2025towards,pan2025self} leveraging RL-style rewards for aesthetic preference alignment or semantic consistency. 
But these methods focus primarily on generative synthesis rather than image restoration, and their reward formulations are not designed for the physical or clinical characteristics of surgical imagery. 

\noindent\textbf{Summary.}
Traditional desmoking methods depend on handcrafted priors that fail under surgical illumination, learning-based restorers require paired supervision that is rarely available. At the same time, RL-driven diffusion methods target generative preference alignment rather than restoration quality. 
These limitations motivate the development of PhySe-RPO, which reformulates diffusion restoration as a stochastic policy optimization problem and introduces physics-guided and semantic rewards tailored for surgical smoke removal.

\section{Methods}
\begin{figure*}[t!]
    \centering 
    \includegraphics[width=\textwidth,keepaspectratio]{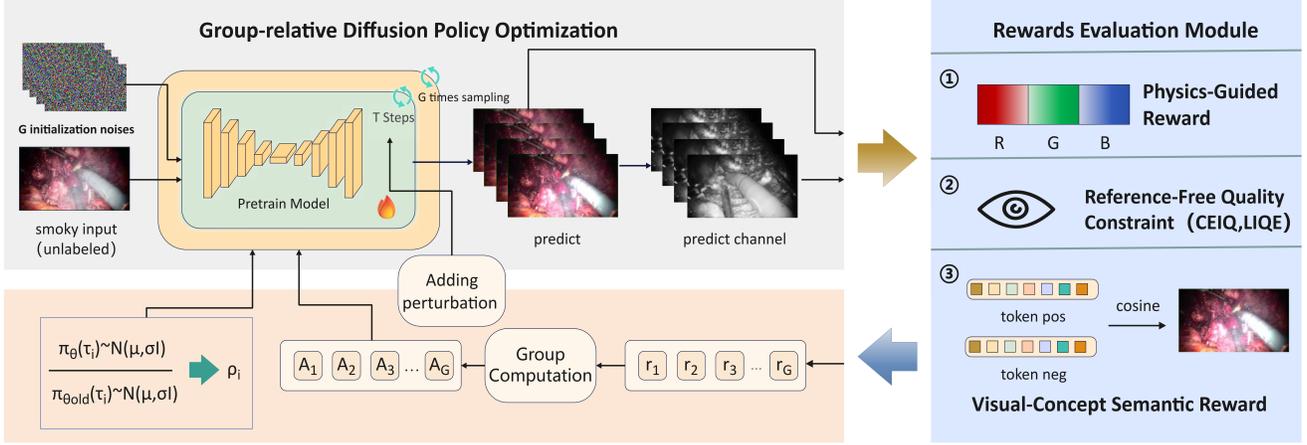} 
    \caption{\textbf{Overview of the PhySe-RPO framework.}  
PhySe-RPO refines the pretrained diffusion model through Group-relative Diffusion Policy Optimization,  
where multiple stochastic trajectories are sampled and optimized using physics-guided color priors, perceptual quality metrics, and semantic rewards, achieving physically consistent and clinically interpretable surgical smoke removal.}
\vspace{-10pt} 
\label{fig:framework2} 
\end{figure*}
We formulate surgical smoke removal as a stochastic diffusion policy learning problem that enables reward-guided refinement without paired supervision.  
Our framework introduces physics-based color priors, reference-free perceptual constraints, and visual-concept semantic rewards to jointly enforce physical realism and semantic fidelity.  
Through this unified design, the model achieves physically grounded, perceptually consistent, and semantically coherent restoration of real robotic surgical videos. The whole framework is presented in Figure~\ref{fig:framework2}.

\subsection{Group-relative Diffusion Policy Optimization}

Reinforcement learning has achieved remarkable progress in aligning large language models through Group Relative Policy Optimization (GRPO)~\cite{shao2024deepseekmath}, yet its application to image restoration remains underexplored. 
Unlike text generation, which naturally supports one-to-many mappings, diffusion-based image restoration typically follows a deterministic one-to-one correspondence between degraded and clean images. 
Such determinism constrains the model’s exploration capability and limits the diversity of restoration trajectories, particularly under unpaired supervision where ground-truth references are unavailable. 
To overcome this limitation, we reformulate diffusion-based restoration as a stochastic policy learning problem, introducing controlled perturbations into the diffusion process to enable exploration and improve reward sensitivity.

\noindent\textbf{Perturbation-driven Stochastic Sampling.}  
To enable exploration in the diffusion policy space, we inject controlled Gaussian perturbations into the sampling process, transforming the deterministic denoising trajectory into a stochastic one.
Specifically, we randomly initialize multiple Gaussian noise and start to denoise from different perturbed states step by step, which introduces greater randomness and prevents the model from collapsing into a single restoration path. 
Formally, given an input $x$, condition $c$, and diffusion time step $t$, we generate multiple restoration candidates through perturbed diffusion trajectories:
\begin{equation}
    \tilde{x}^{(g)} = \mathcal{F}_\theta(x; \epsilon^{(g)}, c, t), 
    \quad \epsilon^{(g)} \sim \mathcal{N}(0, I), \;
    g = 1,2,\dots,G,
    \label{eq:stochastic_sampling}
\end{equation}
where $\mathcal{F}_\theta$ denotes the diffusion process parameterized by $\theta$, $G$ represents the number of random samples, and $\epsilon^{(g)}$ represents the $g$-th Gaussian perturbation.
Each perturbed path yields a physically plausible yet perceptually diverse restoration candidate $\tilde{x}^{(g)}$, forming a group $\{\tilde{x}^{(g)}\}_{g=1}^G$ for subsequent policy evaluation.\\
Meanwhile, we also add random noise perturbations to each step of the denoising process in diffusion, further enhancing the exploration of random strategies:
\begin{equation}
x_{t-1} = \mu_{\theta}(x_t, t) + \sigma_t \epsilon, \quad \epsilon \sim \mathcal{N}(0, I)
\end{equation}
This perturbation-driven stochastic sampling transforms deterministic diffusion into a stochastic policy exploration procedure. 
By producing multiple semantically consistent trajectories, the model can evaluate diverse outcomes under identical inputs, allowing reward-based discrimination and stable optimization without paired supervision. 

\noindent\textbf{Stochastic Policy Optimization.}  
Building on the stochastic sampling strategy introduced above, we further extend the diffusion model into a learnable policy optimized via group-relative reinforcement learning~\cite{shao2024deepseekmath}, specifically adapted for diffusion-based image restoration.  
While GRPO has demonstrated strong stability and sample efficiency in aligning large language models, directly applying it to image restoration is non-trivial due to the continuous nature of diffusion trajectories and the absence of discrete actions.  
Unlike GRPO, which optimizes discrete token-level probabilities, our method treats the diffusion process as a stochastic policy that generates denoising trajectories under Gaussian perturbations, thereby extending relative policy optimization to high-dimensional image generation.

Given a group of $G$ restoration trajectories with corresponding rewards $\{r_i\}_{i=1}^G$, we compute the normalized relative advantage as:
\begin{equation}
\hat{A}_i = \frac{r_i - \bar{r}}{\sigma_r},
\label{eq:grpo_adv}
\end{equation}
where $\bar{r}$ and $\sigma_r$ denote the mean and standard deviation of the rewards within the group.  
This normalization eliminates the dependence on an explicit critic network, stabilizing optimization through group-wise variance reduction.

The diffusion model acts as a stochastic policy $\pi_\theta$ that produces denoising trajectories parameterized by $\theta$.  
The policy is optimized using a clipped surrogate objective:
\begin{equation}
\mathcal{L}_{\text{RPO}}(\theta)
= \mathbb{E}\!\left[\tfrac{1}{G}\!\sum_{i=1}^{G}
\min\!\big(\rho_i\hat{A}_i,\,
\text{clip}(\rho_i,1\!-\!\epsilon,1\!+\!\epsilon)\hat{A}_i\big)\right],
\label{eq:rpo_loss}
\end{equation}
where:
\begin{equation}
\begin{aligned}
\rho_i(\theta) 
&= \frac{\pi_\theta(\tau_i)}{\pi_{\theta_{\text{old}}}(\tau_i)} 
\\[3pt]
&= 
\exp\!\left[
-\frac{1}{2\sigma_i^2}
\left(
\|\tau_i - \mu_\theta\|^2 
- 
\|\tau_{i_{\text{old}}} - \mu_{\theta_{\text{old}}}\|^2
\right)
\right].
\end{aligned}
\label{eq:ratio}
\end{equation}
Here, $\pi_\theta(\tau_i)$ and $\pi_{\theta_{\text{old}}}(\tau_i)$ represent the likelihoods of diffusion trajectories under the current and previous policies, 
$\mu_\theta$ and $\mu_{\theta_{\text{old}}}$ denote predicted means, 
$\sigma_i^2$ is the variance, and $\epsilon$ is the clipping threshold that constrains policy updates.

To prevent policy drift and maintain the fidelity of the pretrained diffusion prior, 
we adopt a KL-regularization term with respect to a frozen reference model $\pi_{\theta_{\text{ref}}}$:
\begin{equation}
\mathcal{L}_{\text{total}} =
\mathcal{L}_{\text{RPO}} + 
\lambda_{\text{KL}} D_{\text{KL}}\big(\pi_\theta \,\|\, \pi_{\theta_{\text{ref}}}\big).
\label{eq:total_objective}
\end{equation}
This design extends GRPO into the diffusion based restoration domain, 
enabling critic-free and variance-reduced policy optimization that integrates stochastic exploration into the image restoration process.  
Through iterative refinement guided by relative rewards, the model learns a stable restoration policy well-suited to unpaired surgical data.

\subsection{Physics-Guided Reward via Color Priors}

Surgical smoke alters light scattering and tissue reflectance, causing spectral imbalance and perceptual color distortion in endoscopic images. 
To restore physically plausible appearance, we introduce a \textit{physics-guided reward} based on color priors that model inter- and intra-channel relationships of RGB components. 
These priors explicitly regularize illumination and color consistency, serving as physically interpretable constraints for diffusion-based desmoking under unpaired supervision.  
For each color channel $c \in \{R, G, B\}$, we define $\mu_c$, $\sigma_c$, and $\text{grad}_c$ 
as the mean intensity, standard deviation, and average gradient magnitude, which respectively characterize brightness, contrast, and edge strength of the channel.  
These statistics jointly provide a compact photometric representation of smoke-induced degradation and form the basis of our color-based reward formulation.

\noindent\textbf{Inter-channel Prior.}  
Driven by the spectral asymmetry inherent in real surgical imaging~\cite{xia2024new}, 
this prior preserves natural inter-channel relationships and mitigates color shifts caused by smoke scattering or over-enhancement.  
In endoscopic scenes, the red–green and red–blue channel differences are generally larger than the green–blue difference, reflecting illumination bias and tissue reflectance characteristics.  
To encode this property, we compute the 95th percentile of absolute mean differences between each channel pair as stable reference statistics:
\begin{equation}
\begin{aligned}
\text{MRG} &= P_{95}(|\mu_R - \mu_G|), \\
\text{MRB} &= P_{95}(|\mu_R - \mu_B|), \\
\text{MGB} &= P_{95}(|\mu_G - \mu_B|),
\end{aligned}
\end{equation}
where $P_{95}(\cdot)$ denotes the 95th percentile estimated from real surgical data.  
Deviations from these natural relationships are penalized by:
\begin{equation}
\begin{aligned}
L_{RG} &= \max(0, \text{MRG} - |\mu_R - \mu_G|), \\
L_{RB} &= \max(0, \text{MRB} - |\mu_R - \mu_B|), \\
L_{GB} &= \max(0, |\mu_G - \mu_B| - \text{MGB}),
\end{aligned}
\end{equation}
and the corresponding reward term is formulated as:
\begin{equation}
R_A = - (L_{RG} + L_{RB} + L_{GB}).
\end{equation}
This prior enforces physiologically consistent inter-channel color contrast, 
reducing hue bias and promoting perceptually balanced reconstruction.  
By embedding this constraint into the reward, the diffusion policy learns to restore color distributions that align with real tissue reflectance and illumination statistics.\\

\noindent\textbf{Intra-channel Prior.}  
Complementary to the global inter-channel constraint, this intra-channel prior focuses on per-channel stability to prevent over-correction of the relatively stable red channel while enhancing contrast recovery in the smoke-degraded green and blue channels.  
Absolute deviations between input and restored images are computed as:
\begin{equation}
\begin{aligned}
\Delta \mu_c   &= |\mu_c^{\text{predict}} - \mu_c^{\text{in}}|, \\
\Delta \sigma_c &= |\sigma_c^{\text{predict}} - \sigma_c^{\text{in}}|, \\
\Delta \text{grad}_c &= |\text{grad}_c^{\text{predict}} - \text{grad}_c^{\text{in}}|.
\end{aligned}
\end{equation}
The reward encourages stronger recovery in the green and blue channels, 
while constraining excessive correction in red to preserve global color fidelity:
\begin{equation}
\begin{aligned}
R_B &= 
\frac{\Delta \mu_G + \Delta \mu_B}{2} - \Delta \mu_R 
+ \frac{\Delta \sigma_G + \Delta \sigma_B}{2} - \Delta \sigma_R \\
&\quad + \frac{\Delta \text{grad}_G + \Delta \text{grad}_B}{2} - \Delta \text{grad}_R.
\end{aligned}
\end{equation}
Therefore, we combine them and call it:
\begin{equation}
\begin{aligned}
R_{PG} &= R_A+R_B
\end{aligned}
\end{equation}
Together, $R_A$ and $R_B$ introduce physically grounded constraints that regulate both inter-channel harmony and intra-channel stability, 
serving as key components of the physics-guided reward in PhySe-RPO. 
They guide the diffusion model toward desmoking results that are visually natural, spectrally balanced, and physiologically interpretable. 
\label{sec:intro}
\begin{figure}[t] 
    \centering 
    \includegraphics[width=\columnwidth,keepaspectratio]{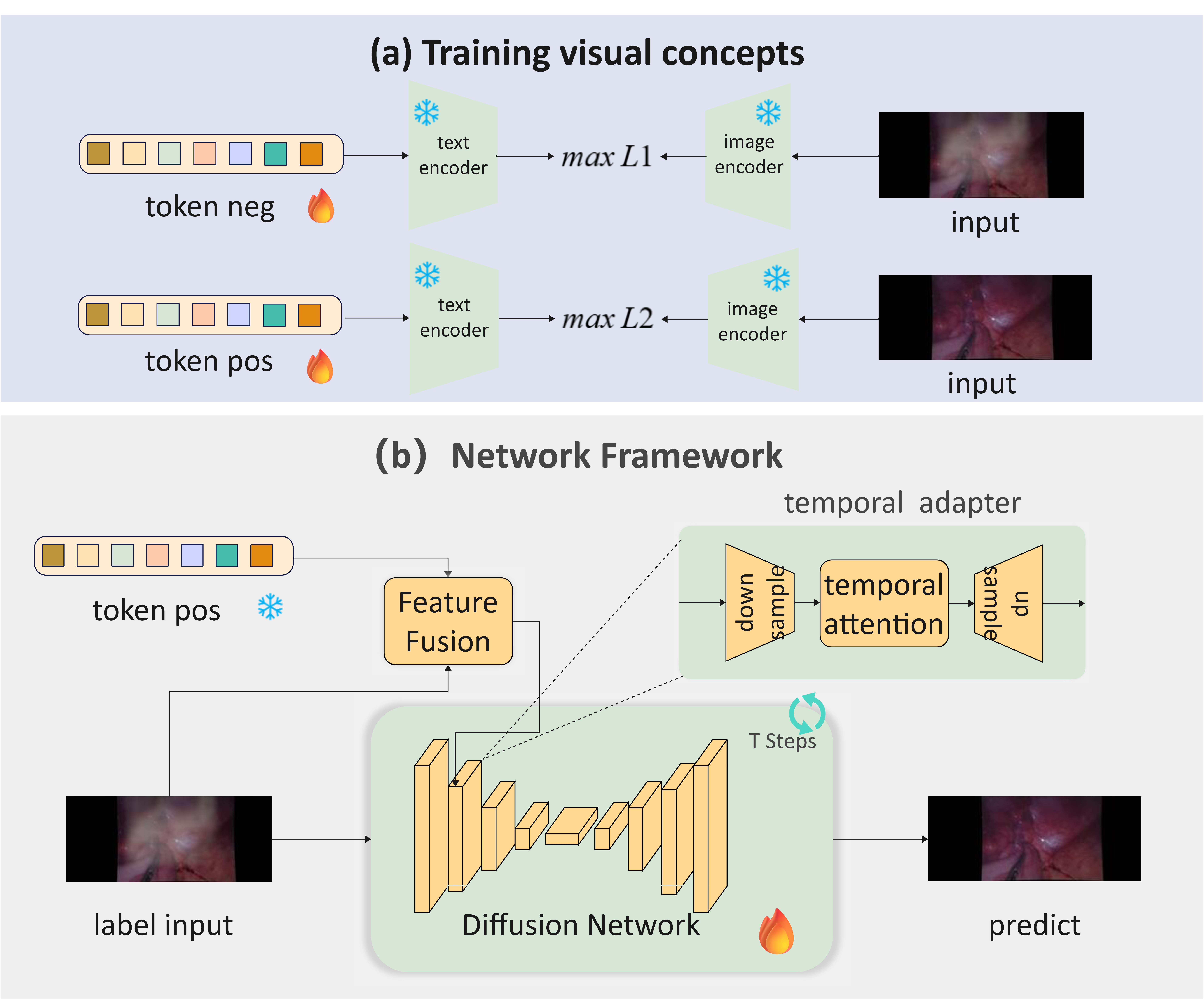}
    \caption{ \textbf{Visual-Concept Integration into Diffusion.}  
(a) Learnable visual concepts are trained via contrastive learning to differentiate ``clear’’ and ``smoky’’ concepts in the semantic space.  
(b) The learned tokens are integrated into the diffusion backbone through multimodal fusion and temporal adaptation to guide semantically consistent desmoking.
} 
    \label{fig:framework}

\end{figure}
\subsection{Visual-Concept Semantic Reward}

While physics-based priors regularize low-level color and illumination consistency, they cannot ensure that the restored image preserves the semantic integrity of the surgical scene.  
To provide high-level semantic guidance, we introduce a \textit{Visual-Concept Semantic Reward} that evaluates whether the generated image aligns with the learned ``clear’’ visual concept rather than the ``smoky’’ one.  
Unlike language-driven rewards, this formulation operates purely in the vision–semantic space, enabling the model to leverage CLIP’s multimodal representations without relying on explicit textual supervision.

\noindent\textbf{Learning Visual Concepts.}  
We first establish domain-specific visual concepts from synthetic paired data by learning CLIP-based visual tokens, as illustrated in Figure~\ref{fig:framework}(a).  
Inspired by Context Optimization (CoOp)~\cite{zhou2022learning}, we introduce learnable positive and negative tokens representing ``clear’’ and ``smoky’’ surgical states, respectively.  
Each token sequence is embedded via a frozen text encoder $\varepsilon_t(\cdot)$ and aligned with image embeddings from a frozen visual encoder $\varepsilon_i(\cdot)$ through cosine similarity:
\begin{equation}
\begin{aligned}
L_{\text{neg}} &= \cos(\varepsilon_t(\text{neg}), \varepsilon_i(\text{LQ})), \\
L_{\text{pos}} &= \cos(\varepsilon_t(\text{pos}), \varepsilon_i(\text{HQ})).
\end{aligned}
\end{equation}
The objective enhances the alignment between clear images and positive tokens while strengthening the association between smoky images and negative tokens:
\begin{equation}
L_{match} = - (L_{\text{neg}} + L_{\text{pos}}).
\end{equation}
Through this contrastive supervision, the tokens evolve into domain-adaptive visual concepts that reside in the same embedding manifold as surgical images.  
This adaptation effectively mitigates the the distributional discrepancy between textual and visual embeddings in CLIP, which becomes pronounced in surgical domains due to domain-specific appearance patterns and semantic contexts absent from natural-image pretraining.

\noindent\textbf{Integrating Visual Concepts into Diffusion.}  
The learned tokens are then incorporated into the diffusion backbone as semantic conditions for desmoking, as shown in Figure~\ref{fig:framework}(b).  
Specifically, the frozen CLIP encoders provide multimodal embeddings, while the learnable tokens are injected into the denoising network via a semantic–visual fusion module based on cross-attention.  
This mechanism allows the diffusion model to integrate high-level semantic priors into the generative process, preserving anatomical structures and contextual coherence during restoration.  
The detailed architecture of the multimodal fusion module is provided in the Appendix.

\noindent\textbf{Concept-Level Reward Formulation.}  
After the visual concepts are learned during pretraining, the frozen CLIP visual encoder together with the learned positive and negative concept serves as a semantic evaluator during reward-based optimization.  
Let $v_I$ denote the embedding of a generated desmoked image, and let $v_T^+$ and $v_T^-$ be the learned ``clear’’ and ``smoky’’ tokens, respectively.  
We define a concept-level semantic reward as the log-probability of $v_I$ being aligned with the ``clear’’ concept:
\begin{equation}
R_{\text{VC}} = 
\log \frac{
\exp(\cos(v_I, v_T^+)/\tau)
}{
\exp(\cos(v_I, v_T^+)/\tau)
+ 
\exp(\cos(v_I, v_T^-)/\tau)
},
\label{eq:reward_sem}
\end{equation}
where $\tau$ is a temperature parameter controlling similarity sharpness.  
This reward encourages generated images to move toward the ``clear’’ semantic manifold while being pushed away from the ``smoky’’ manifold, improving both perceptual clarity and semantic coherence.  
By evaluating restored images directly within the visual embedding space, the framework avoids cross‐modal mismatch and achieves stable, concept‐aligned reinforcement optimization.

\subsection{Reference-Free Quality Constraint}
To provide perceptual guidance under unpaired conditions, we introduce a reference-free quality constraint based on learned image quality assessment (IQA) models. 
Specifically, we employ two complementary IQA metrics: 
CEIQ (Contrastive Explainable Image Quality) and LIQE (Learned Image Quality Evaluator).  
Both are pre-trained on large-scale aesthetic and perceptual datasets~\cite{yan2019no,zhang2023blind} and can infer human-perceived attributes such as clarity, contrast, and naturalness directly from single images.  
Let $\hat{x}$ denote the desmoked image generated by the diffusion model.  
By integrating the IQA scores of $\hat{x}$ into the reward, the diffusion policy is encouraged to produce visually pleasing and artifact-free outputs without relying on paired ground-truth supervision.  
Formally, the reward is defined as:
\begin{equation}
R_{RF} = CEIQ(\hat{x}) + LIQE(\hat{x}).
\label{eq:reward_iqa}
\end{equation}
This perceptual component complements the physical priors by aligning the generation process with human visual perception, ensuring that restored frames remain coherent, and diagnostically reliable across surgical video sequences.

\subsection{Overall}
The overall reward integrates the complementary objectives described above to jointly optimize the desmoking policy.  
The physical priors ($R_{PG}$) constrain color statistics and inter-channel relationships,  
the reference-free quality term ($R_{RF}$) enforces visually coherent and artifact-free restoration,  
and the vision concept reward ($R_{\text{VC}}$) introduces a high-level discriminative signal indicating smoke-free appearance.  
The final composite reward is defined as:
\begin{equation}
R = R_{PG} + R_{RF} + R_{\text{VC}}.
\label{eq:total_reward}
\end{equation}
This unified formulation bridges low-level physical fidelity and high-level semantic perception,  
enabling reinforcement-guided optimization that drives the diffusion model toward realistic, robust, and perceptually consistent desmoking results.
\begin{figure*}[t] 
    \centering 
    \includegraphics[width=\textwidth,keepaspectratio]{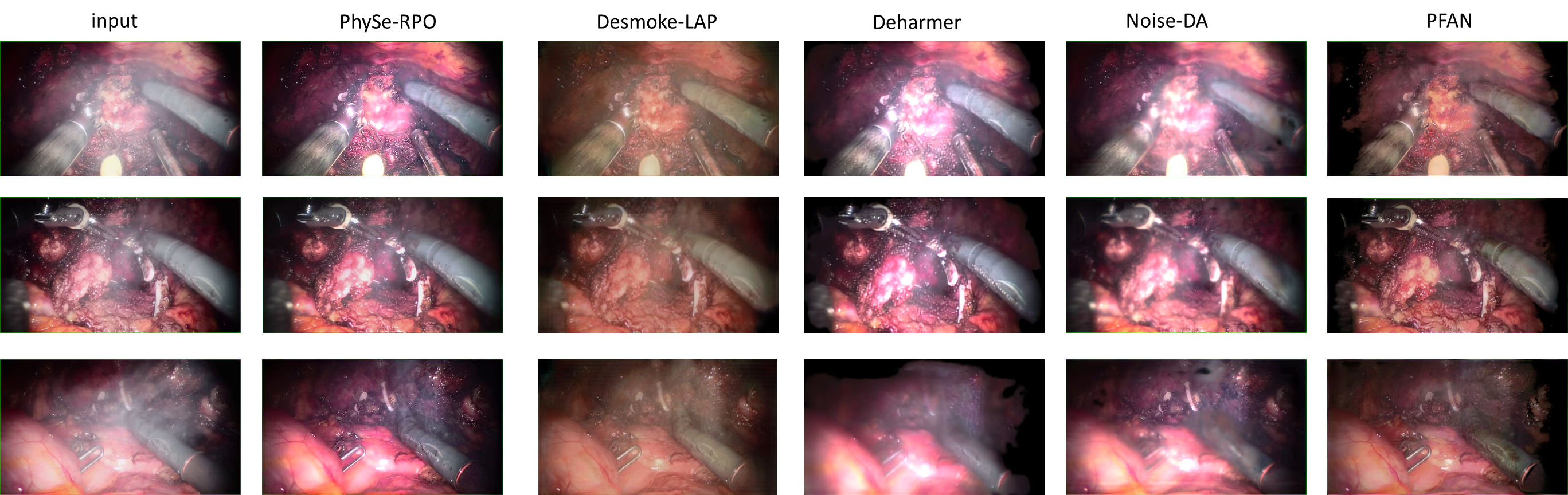}
    \caption{
Qualitative comparison on real-world surgical smoke images. 
Compared with prior desmoking methods, PhySe-RPO produces clearer structures, more natural color restoration, and fewer residual smoke artifacts.
}

    \label{fig:4} 
\end{figure*}

\section{Experiments}
\subsection{Experimental Settings}

We evaluate our method under a pipeline that requires both paired supervised data for cold-start training and unlabeled real surgical data for reward-guided refinement.

\noindent\textbf{Synthetic Paired Dataset.}
To obtain controllable supervision as cold start, we construct a synthetic surgical smoke dataset using Blender with volumetric rendering for realistic smoke dispersion. A total of 2,000 paired smoky–clean images have been generated. Following prior diffusion-based restoration works, 1,600 pairs are used to pretrain the model and learn a structure-preserving de-smoking prior, while the remaining 400 pairs serve as a validation set for evaluation.

\noindent\textbf{Real Surgical Dataset.}
To capture the complexity of real operating environments, we collect a robotic surgical dataset containing 10,000 unlabeled frames from in vivo procedures. These videos cover diverse smoke densities, tissue appearances, and illumination conditions not reproducible through simulation. During PhySe-RPO refinement, the unlabeled images are used to optimize the diffusion policy using physics- and semantics-guided rewards. Additionally, 400 real surgical frames are held out as a test set for evaluating performance on real clinical data.

\noindent\textbf{Implementation Details:}
The diffusion model is trained using four NVIDIA H100 GPUs with the AdamW optimizer $(\beta_1 = 0.9, \beta_2 = 0.99)$. 
The batch size is set to 16, and the number of diffusion time steps $T$ is 100. The group number $G$ is set to 4, and the clipping parameter $\epsilon$ is set to 0.2. The CLIP text and image encoders adopted in this framework are based on ViT-B/32. 

\noindent\textbf{Evaluation Metrics:}
Since no ground-truth reference is available, we employ several no-reference image quality assessment (NR-IQA) metrics to evaluate the real-world performance. The selected indicators include: SSEQ \cite{liu2014no}, MANIQA \cite{yang2022maniqa}, PI \cite{blau2018perception}, FADE \cite{choi2015referenceless}, MUSIQ \cite{ke2021musiq}, IS (Inception Score) \cite{salimans2016improved}, and NIQE \cite{mittal2012making}. These no-reference indicators provide a comprehensive assessment from the perspectives of visual quality, contrast, perceptual realism, and naturalness, reflecting the model’s generalization ability in real-world surgical de-smoking scenarios.

\begin{table*}[t]
\centering
\caption{
Comparison with state-of-the-art desmoking methods on the real-world surgical dataset. PhySe-RPO achieves the best overall performance across no-reference quality metrics.
}
\label{tab:comparison_real}
\begin{tabular}{lccccccc}
\toprule
\textbf{Method} & \textbf{SSEQ↓} & \textbf{MANIQA↑} & \textbf{PI↓} & \textbf{FADE↓} & \textbf{MUSIQ↑} & \textbf{IS↑} & \textbf{NIQE↓} \\
\midrule
DCP~\cite{he2010single} & 24.944 & 0.295 & 3.577 & 0.401 & 49.385 & 2.394 & 5.732 \\
Desmoke\_LAP~\cite{pan2022desmoke} & 32.305 & 0.178 & 5.675 & 0.604 & 38.420 & 2.010 & 6.504 \\
SelfSVD~\cite{wu2024self} & 11.868 & 0.253 & 3.227 & 0.415 & 48.422 & 2.307 & \textbf{4.320} \\
PFAN~\cite{zhang2023progressive} & 30.876 & 0.248 & 3.562 & 0.356 & 46.656 & 2.365 & 5.163 \\
Dehamer~\cite{guo2022image} & 36.741 & 0.134 & 4.713 & 0.516 & 33.995 & 2.469 & 6.349 \\
Tap~\cite{fu2024temporal} & 16.646 & 0.259 & 3.447 & 0.428 & 44.094 & 2.415 & 5.582 \\
LightDiff~\cite{chen2024lightdiff} & 28.624 & 0.164 & 3.964 & 0.547 & 38.251 & 2.523 & 5.155 \\
Noise-DA~\cite{liao2024denoising} & 36.526 & 0.096 & 6.395 & 0.553 & 31.786 & \textbf{3.000} & 6.673 \\
DGFDNet~\cite{zheng2025efficient} & 36.286 & 0.208 & 3.891 & 0.354 & 48.053 & 2.169 & 5.362 \\
PhySe-RPO & \textbf{3.443} & \textbf{0.378} & \textbf{3.125} & \textbf{0.216} & \textbf{54.911} & \underline{2.797} & \underline{4.608} \\
\bottomrule
\end{tabular}
\end{table*}
\begin{table*}[h]
\centering
\caption{Ablation results of PhySe-RPO on the real-world dataset.}
\label{tab:ablation_stage2}
\begin{tabular}{lccccccc}
\toprule
\textbf{Method} & \textbf{SSEQ↓} & \textbf{MANIQA↑} & \textbf{PI↓} & \textbf{FADE↓} & \textbf{MUSIQ↑} & \textbf{IS↑} & \textbf{NIQE↓} \\
\midrule
Baseline & 7.149 & 0.332 & 4.111 & 0.420 & 45.586 & 2.613 & 5.813 \\
Baseline+$R_{RF}$ & 5.018  & 0.318 & 3.288 & 0.360 & 49.582 & 2.730 & 4.944 \\
Baseline+$R_{RF}$+$R_{VC}$ & 4.771 & 0.368 & 3.235 & 0.246 & 53.818 & 2.764 & 4.752 \\
Baseline+$R_{RF}$+$R_{VC}$+$R_{PG}$ & \textbf{3.443} & \textbf{0.378} & \textbf{3.125} & \textbf{0.216} & \textbf{54.911} & \textbf{2.797} & \textbf{4.608} \\
\bottomrule
\end{tabular}
\end{table*}

\subsection{Experimental Results}
We adopt the diffusion model architecture proposed in~\cite{luo2023controlling} as the backbone of our framework, providing a stable and expressive generative foundation for modeling the surgical desmoking process. To comprehensively evaluate the effectiveness of the proposed PhySe-RPO, we compare it against a wide range of classical, learning-based, and diffusion-based methods. Specifically, we include the prior-based DCP~\cite{he2010single}, learning-based networks such as Desmoke\_LAP~\cite{pan2022desmoke}, SelfSVD~\cite{wu2024self}, and PFAN~\cite{zhang2023progressive}, as well as recent diffusion- and transformer-based techniques including Dehamer~\cite{guo2022image}, TAP~\cite{fu2024temporal}, LightDiff~\cite{chen2024lightdiff}, Noise-DA~\cite{liao2024denoising}, and DGFDNet~\cite{zheng2025efficient}.

\noindent\textbf{Evaluation on Unlabeled Real Surgical Data.}
We first evaluate the proposed framework on our self-constructed unlabeled robotic surgical dataset. As shown in Table~\ref{tab:comparison_real}, our PhySe-RPO consistently surpasses all competing methods across nearly all reference-free image quality metrics. The model achieves the lowest SSEQ (3.443), PI (3.125), and FADE (0.216), indicating improved structural coherence and contrast recovery, while also obtaining the highest MANIQA (0.378) and MUSIQ (54.911), reflecting enhanced perceptual quality. These results show that PhySe-RPO effectively leverages physics and semantics guided rewards to refine the diffusion model on real surgical data, producing restorations that are both visually natural and physically plausible despite the lack of paired supervision.

\noindent\textbf{Evaluation on Public Paired Dataset.}  
To further assess the quantitative restoration capability of our model, we evaluate PhySe-RPO on a publicly available Surgical Paired Dataset proposed by~\cite{xia2024new}, which provides clean references for objective comparison and enables reliable benchmarking of restoration performance across different surgical scenes.
As summarized in Table~\ref{tab:comparison_public}, our method achieves the highest PSNR (21.03~dB) and lowest CIEDE-2000 (7.65), outperforming state-of-the-art approaches such as DehazeFormer~\cite{song2023vision} and Fog-Removal~\cite{jin2022structure}.
Although these pixel-level metrics have inherent limitations in capturing perceptual or semantic quality, PhySe-RPO consistently attains the best results across both reference-based and reference-free evaluations, further demonstrating its robustness, stability, and strong generalization capability under diverse and challenging surgical imaging conditions.

\noindent\textbf{Qualitative Comparison.}  
Figure~\ref{fig:4} provides visual comparisons on real-world surgical smoke scenes. Competing methods, including Desmoke\_LAP, Dehamer, Noise-DA, and PFAN, often fail to completely remove dense smoke or introduce over-saturation and halo artifacts that degrade visual consistency. In contrast, our approach produces clearer and more realistic surgical views with balanced illumination and faithful tissue color reproduction. 
Fine anatomical structures, such as vessel edges and instrument boundaries, are well preserved, confirming that PhySe-RPO achieves physically consistent and clinically interpretable restoration across diverse surgical environments.

\subsection{Ablation Study}
To further evaluate the contribution of each reward component in the PhySe-RPO framework, we conduct ablation experiments based on the pre-trained diffusion model before policy optimization, which serves as the baseline. 
We sequentially incorporate the proposed reward terms: Reference-Free Quality Constraint, Visual-Concept Semantic Reward, and Physics-Guided Reward, to assess their individual and joint effects. 
As shown in Table~\ref{tab:ablation_stage2}, the complete configuration achieves the best trade-off between perceptual naturalness and semantic fidelity, demonstrating that the multi-reward design of PhySe-RPO effectively guides diffusion refinement toward high-quality, visually coherent surgical view restoration. 
Moreover, the progressive performance gains reveal that each reward contributes complementary advantages, confirming that jointly optimizing physical consistency, semantic correctness, and perceptual realism is essential for robust smoke removal under unpaired surgical scenarios.

\begin{table}[t]
\centering
\caption{Comparison of different semantic reward strategies on the real-world surgical dataset. The proposed Visual-Concept Reward yields the best perceptual quality and lowest distortion.}
\label{tab:ablation_vc}
\setlength{\tabcolsep}{2.5pt} 
\resizebox{\columnwidth}{!}{
\begin{tabular}{lccccccc}
\toprule
\footnotesize \textbf{Method} & \footnotesize \textbf{SSEQ↓} & \footnotesize \textbf{MANIQA↑} & \footnotesize \textbf{PI↓} & \footnotesize \textbf{FADE↓} & \footnotesize \textbf{MUSIQ↑} & \footnotesize \textbf{IS↑} & \footnotesize \textbf{NIQE↓}\\
\midrule
w/o R$_{\text{VC}}$ & 5.264 & 0.320 & 3.664 & 0.278 & 51.517 & 2.639 & 5.157\\
Text & 4.304 & 0.308 & 3.347 & 0.244 & 53.269 & 2.773 & 4.867 \\
\textbf{Full} & \textbf{3.443} & \textbf{0.378} & \textbf{3.125} & \textbf{0.216}& \textbf{54.911} & \textbf{2.797} & \textbf{4.608} \\
\bottomrule
\end{tabular}
}
\end{table}
\begin{figure}[t]
    \centering
    \includegraphics[width=\linewidth]{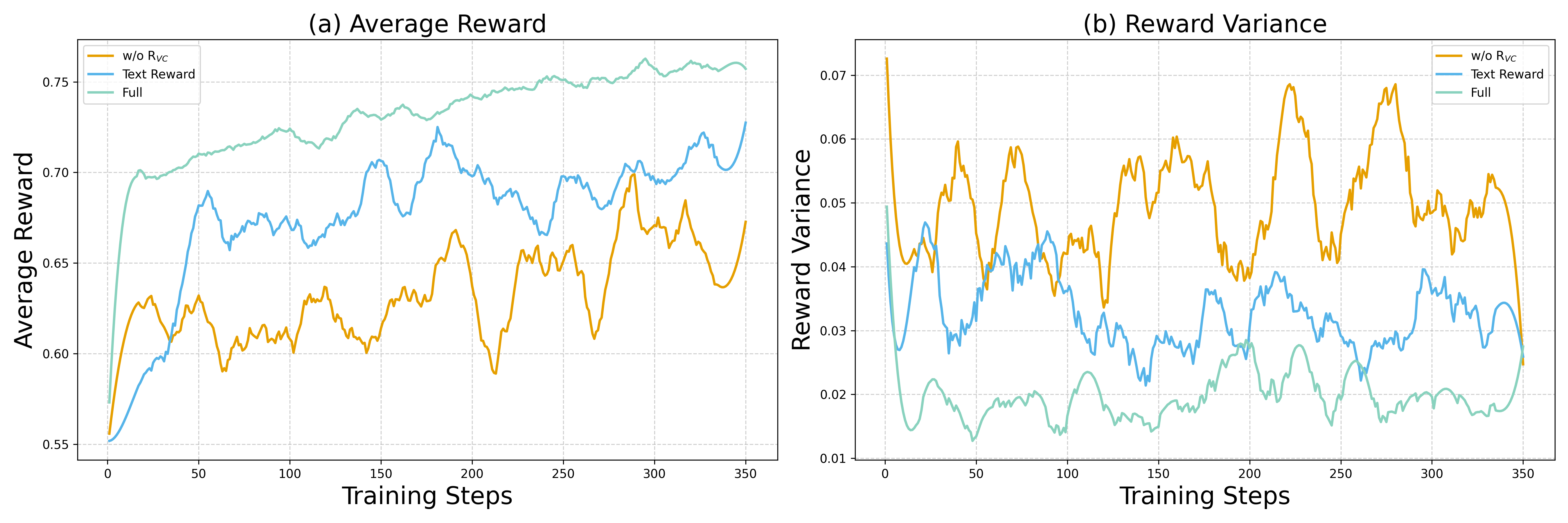}
    \caption{
    \textbf{Reward convergence analysis.} 
Average reward (a) and reward variance (b) under different semantic reward settings. The \textbf{Full} model converges faster with lower variance than \emph{Text-Reward} and \emph{w/o $R_{\text{VC}}$}.}
    \label{fig:reward_curve}
\end{figure}

\noindent\textbf{Why Visual-Concept Reward?} We validate the necessity of the proposed visual-concept semantic reward with two comparisons. First, Table~\ref{tab:ablation_vc} reports results for three variants: the full PhySe-RPO, a version without semantic reward (\emph{w/o VC-Reward}), and a text-prompt reward baseline (\emph{Text-Reward}). Removing or replacing the visual-concept reward consistently degrades performance, demonstrating that concept-level alignment provides essential perceptual and structural guidance. Second, we analyze optimization dynamics. As shown in Figure~\ref{fig:reward_curve}, the \emph{Text-Reward} variant shows higher reward variance and slower convergence due to the cross-modal gap between text embeddings and surgical image distributions. In contrast, the visual-concept reward operates purely in the visual embedding space, yielding lower variance, and faster convergence. These results confirm that domain-adaptive visual concepts offer a more stable and semantically aligned supervision signal than text-based rewards, improving both optimization stability and final perceptual quality.
\subsection{Downstream Validation.}
We further validated the practical value of our method in downstream segmentation~\cite{xue2025td} tasks using MedSAM and AMNCutter~\cite{ma2024segment,sheng2025amncutter}. 
As shown in Table~\ref{tab:comparison_segmentation}, applying our model as a preprocessing step consistently improves IoU and Dice scores, confirming its effectiveness in enhancing subsequent surgical scene understanding.
\begin{table}[h]
\centering
\caption{Comparison of image desmoking methods on the Public Surgical Paired Dataset.}
\label{tab:comparison_public}
\begin{tabular}{l@{\hspace{0pt}}cccc}
\toprule
\textbf{Method} & \textbf{PSNR↑}  & \textbf{CIEDE-2000↓}  \\
\midrule
Desmoke\_LAP\cite{pan2022desmoke} & 17.97 & 11.70 \\
DLSI\cite{salazar2020desmoking} & 19.95 & 7.94  \\
MS-CycleGAN\cite{su2023multi} & 19.44 & 8.67  \\
GSR\cite{sidorov2020generative} & 19.74 & 8.42  \\
DehazeFormer\cite{song2023vision} & 20.80 & 8.27  \\
Fog-Removal\cite{jin2022structure} & 20.29 & 9.45  \\
Var-desmoke\cite{wang2018variational} & 18.36 & 9.37  \\
Vison-defogging\cite{luo2017vision} & 18.77 & 9.25  \\
PhySe-RPO     & \textbf{21.03} & \textbf{7.65} \\
\bottomrule
\end{tabular}
\end{table}
\begin{table}[h]
\centering
\caption{Results of application in the segmentation domain.}
\label{tab:comparison_segmentation}
\begin{tabular*}{\linewidth}{@{\extracolsep{\fill}}lcc}
\toprule
\textbf{Method} & \textbf{IoU↑}  & \textbf{Dice↑}  \\
\midrule
MedSAM & 0.5807 & 0.6772 \\
MedSAM+ours & \textbf{0.5879} & \textbf{0.6834} \\
AMNCutter & 0.7087 & 0.8252 \\
AMNCutter+ours & \textbf{0.7179} & \textbf{0.8347} \\
\bottomrule
\end{tabular*}

\end{table}

\section{Conclusion}
In this work, we present PhySe-RPO, a diffusion-based framework for surgical smoke removal with physics- and semantics-guided relative policy optimization. By reformulating diffusion as a group-relative stochastic policy, the model enables exploration-driven refinement under unpaired surgical conditions. 
A physics-guided reward based on color priors maintains illumination stability and chromatic consistency, while a visual-concept semantic reward preserves anatomical structure and perceptual realism. 
Experiments on unlabeled surgical videos and public paired datasets demonstrate superior visual quality and clinically interpretable restoration.
\section*{Acknowledgement}
This work was supported by the National Natural Science Foundation of China (62401143), the State Key Project of Research and Development Plan (2024YFF1206703), the Natural Science Foundation of
Jiangsu Province (BK20241301), and the Big Data Computing Center of Southeast University.

{
    \small
    \bibliographystyle{ieeenat_fullname}
    \bibliography{main}
}
\clearpage
\setcounter{page}{1}
\maketitlesupplementary

\section{Additional Experimental Results}
\label{sec:rationale}
\subsection{Results on Synthetic Datasets}
To verify that our model has basic desmoking capabilities at cold start, we tested it on a synthetic dataset. 
The experimental results are shown in Table~\ref{tab:comparison},our method achieves the best overall performance across all four evaluation metrics, demonstrating strong desmoking ability even at cold start.
In terms of distortion-oriented metrics, our approach attains the highest PSNR (35.2552 dB) and SSIM (0.9776), indicating that the restored images are closest to the ground truth in both pixel accuracy and structural fidelity.
For perceptual quality, our method also achieves the lowest LPIPS (0.0227), substantially outperforming recent deep models including PFAN, Dehamer, and LightDiff.
In addition, our model attains the best FID (30.5633) among all competitors, which is nearly 40–90 points lower than traditional and GAN-based baselines, indicating that the generated images are not only perceptually plausible but also distributionally closer to clean images.

\begin{figure}[h] 
    \centering 
    \includegraphics[width=\linewidth,keepaspectratio]{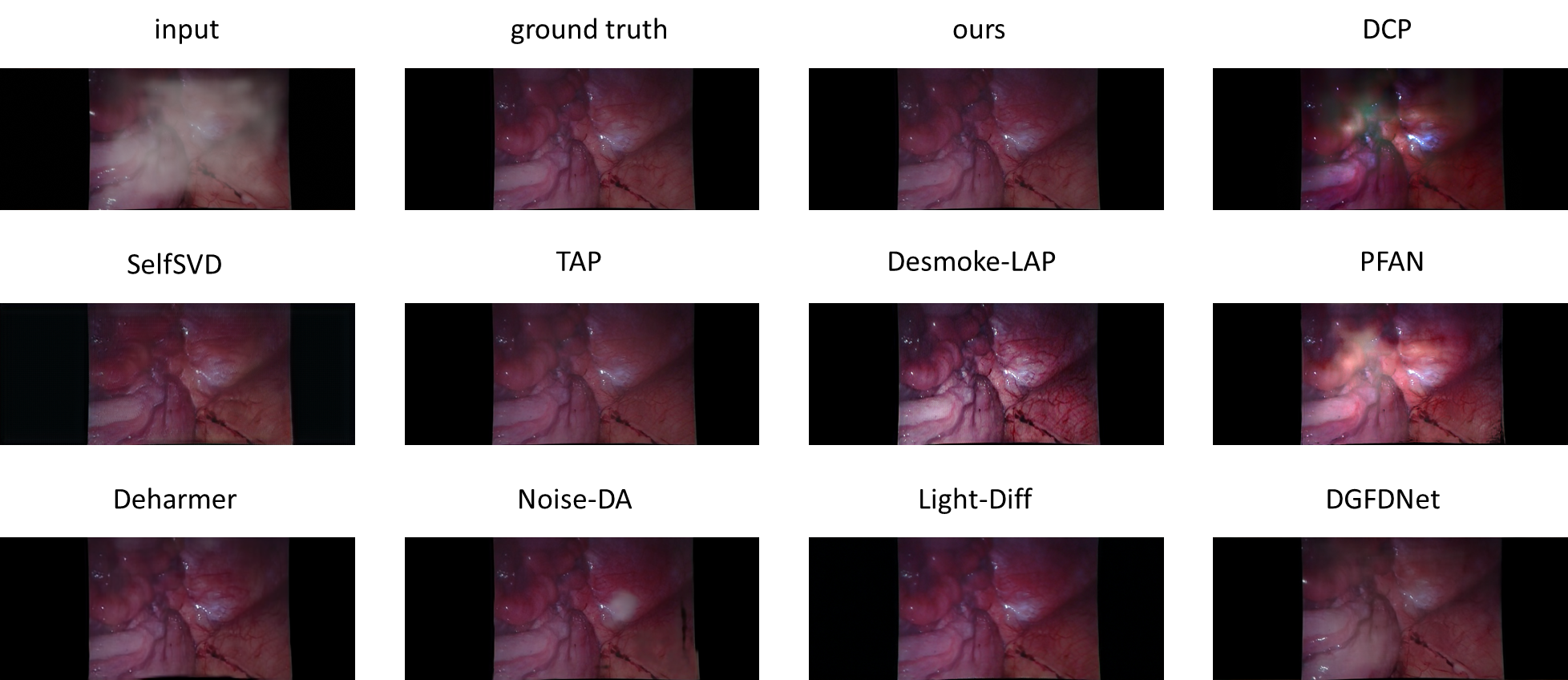} 
    \caption{Qualitative comparison on synthetic images,compared to other methods, PhySe-RPO exhibits better desmoking capabilities in the cold start phase. } 
    \label{fig:6} 
    \vspace{-10pt} 
\end{figure}

\begin{table}[h]
\centering
\caption{Comparison of image desmoking methods on the synthetic dataset in cold start.}
\label{tab:comparison}
\begin{tabular}{l@{\hspace{0pt}}cccc}
\toprule
\footnotesize \textbf{Method} &\footnotesize \textbf{PSNR↑} &\footnotesize \textbf{SSIM↑} &\footnotesize \textbf{LPIPS↓} &\footnotesize \textbf{FID↓} \\
\midrule
\footnotesize DCP\cite{he2010single} &\footnotesize 29.0136 &\footnotesize 0.8529 &\footnotesize 0.1758 &\footnotesize 121.6056 \\
\footnotesize Desmoke\_LAP\cite{pan2022desmoke} &\footnotesize 30.9505 &\footnotesize 0.9144 &\footnotesize 0.0975 &\footnotesize 81.8183 \\
\footnotesize SelfSVD\cite{wu2024self} &\footnotesize 31.2658 &\footnotesize 0.8780 &\footnotesize0.1237 &\footnotesize 150.4745 \\
\footnotesize PFAN\cite{zhang2023progressive} &\footnotesize 30.5614 &\footnotesize 0.9030 &\footnotesize 0.1068 &\footnotesize 111.8937 \\
\footnotesize Dehamer\cite{guo2022image} &\footnotesize 31.6271 &\footnotesize 0.9627 &\footnotesize 0.0406 &\footnotesize 67.8821 \\
\footnotesize Tap\cite{fu2024temporal} &\footnotesize 31.5969 &\footnotesize 0.9622 &\footnotesize 0.0472 &\footnotesize 74.9391 \\
\footnotesize LightDiff\cite{chen2024lightdiff} &\footnotesize 30.5844 &\footnotesize 0.9431 &\footnotesize 0.0874 &\footnotesize 49.9105 \\
\footnotesize Noise-DA\cite{liao2024denoising} &\footnotesize 31.2110 &\footnotesize 0.9568 &\footnotesize 0.0505 &\footnotesize 66.1669 \\
\footnotesize DGFDNet\cite{zheng2025efficient} &\footnotesize 30.7754 &\footnotesize 0.9677 &\footnotesize 0.0452 &\footnotesize 46.5128 \\
\footnotesize Ours     &\footnotesize \textbf{35.2552} &\footnotesize \textbf{0.9776} &\footnotesize \textbf{0.0227} &\footnotesize \textbf{30.5633} \\
\bottomrule
\end{tabular}
\end{table}
In addition to the quantitative results, qualitative comparisons on the synthetic dataset are presented in Figure~\ref{fig:6}. 
Classical prior-based methods, such as DCP~\cite{he2010single}, tend to produce over-saturated colors and residual haze, while learning-based approaches like PFAN~\cite{zhang2023progressive} and Dehamer~\cite{guo2022image} achieve cleaner results but often suffer from texture loss or halo artifacts around structural boundaries. 
Diffusion-based methods (e.g., LightDiff~\cite{chen2024lightdiff}) restore global illumination effectively yet sometimes generate overly smooth surfaces due to weak structural regularization. 
In contrast, our method reconstructs fine anatomical textures and preserves surgical structures more faithfully.
\subsection{Ablation Study in cold start}
To further understand the contribution of each component in our framework, we conducted ablation experiments focusing on the Semantic Feature Fusion (SF) and the proposed Temporal Adapter (TA).
As shown in Table~\ref{tab:ablation_stage1}, both SF and the TA provide clear benefits under the cold-start setting. 
Adding SF yields a notable gain over the baseline , showing that semantic cues effectively guide the diffusion process toward structure-preserving restoration. TA also improves performance by enhancing cross-step consistency. 
Combining both modules leads to the best results for PSNR(35.2552 dB) and SSIM(0.9776), demonstrating their complementary roles in strengthening the model’s early-stage desmoking capability.
\begin{minipage}[t!]{0.48\textwidth}
\centering
\captionof{table}{Ablation results of cold start on the synthetic dataset.}
\label{tab:ablation_stage1}
\begin{tabular}{lcc}
\toprule
\textbf{Model} & \textbf{PSNR↑} & \textbf{SSIM↑} \\
\midrule
Baseline & 34.0958 & 0.9748  \\
Baseline+SF & 34.9618 & 0.9764  \\
Baseline+TA & 34.2361 & 0.9752  \\
Ours(Baseline+SF+TA) & \textbf{35.2552} & \textbf{0.9776} \\
\bottomrule
\end{tabular}
\end{minipage}
\subsection{Ablation on the Physics-Guided Reward Weight}
\begin{figure}[h] 
    \centering 
    \includegraphics[width=\linewidth,keepaspectratio]{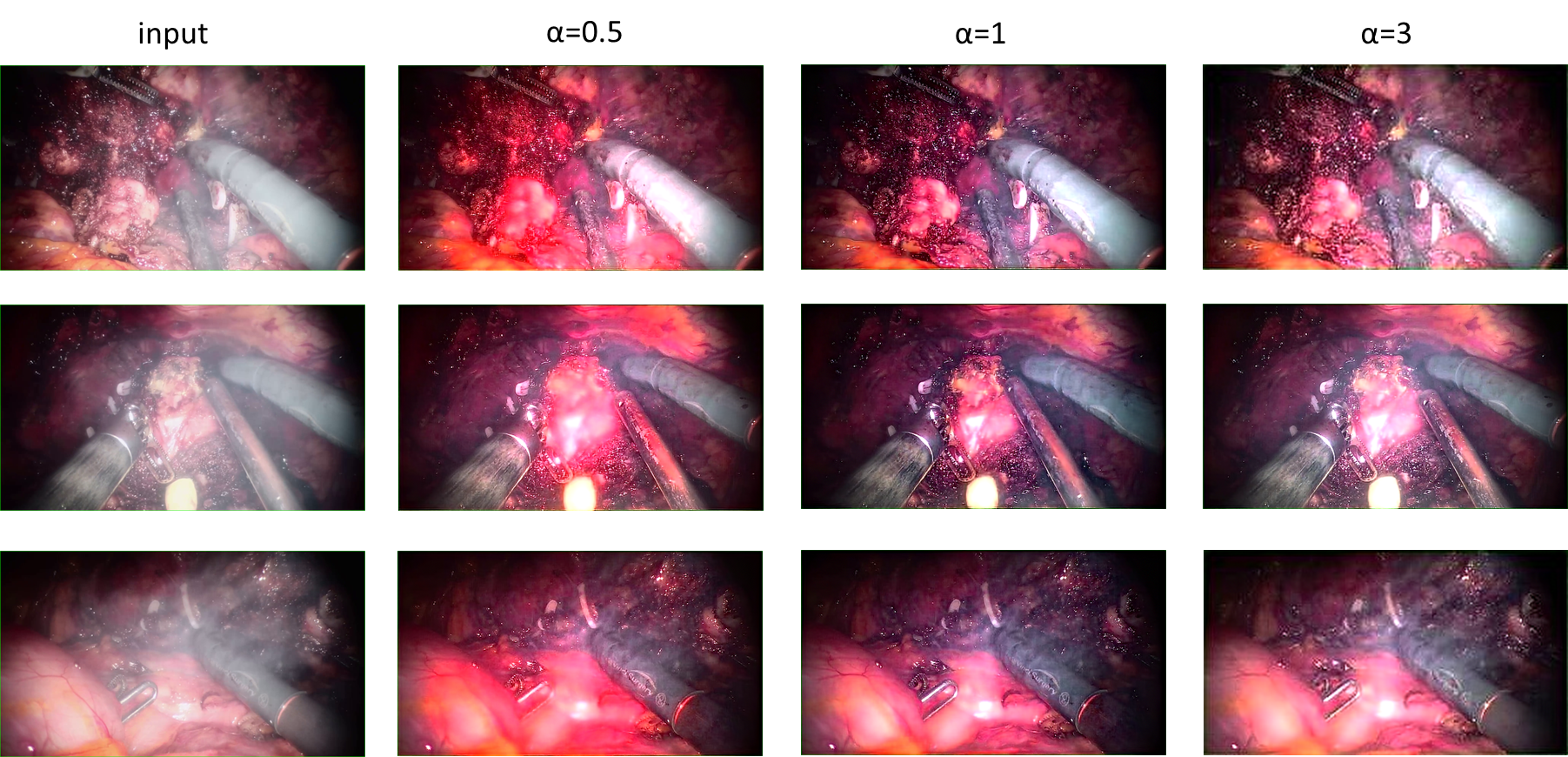} 
    \caption{Output results for different $\alpha$ values} 
    \label{fig:7} 
    \vspace{-10pt}  
\end{figure}

To analyze the influence of the physics-guided term, we reformulate the overall reward as:
\begin{equation}
    R = \alpha R_{PG} + R_{RF} + R_{VC},
\end{equation}
where $\alpha$ controls the strength of the physics-guided prior.
As shown in Figure ~\ref{fig:7},we observe that setting $\alpha$=1 overemphasizes the color prior and leads to reddish artifacts in the desmoked outputs. 
This effect arises because an excessively strong physics-guided constraint encourages over-correction of illumination imbalance, unintentionally amplifying the red channel in surgical scenes.
By changing $\alpha$, the model achieves more natural color reproduction while preserving dehazing performance. 
The results demonstrate that properly tuning $\alpha$ is essential to avoid color shifts and maintain visually balanced reconstruction quality.
\subsection{Hyperparameter experiments}
We further investigate two key hyperparameters in our framework: the length of the learning token embedding and the number of groups $G$ used in the PhySe-RPO.
The learning token length determines the expressiveness of the learned visual concepts. 
As shown in Table~\ref{tab:ablation_token}, varying the token length has only a marginal impact on performance. 
Shorter embeddings already capture sufficient semantic cues to guide the diffusion process, while increasing the token length yields only slight improvements in PSNR and SSIM. 
This indicates that the model is relatively robust to the choice of token dimension, and the semantic concepts can be effectively encoded even with compact embeddings. 
Therefore, a moderate token length provides a good balance between representation capacity and computational cost, without significantly affecting overall performance.

\begin{table}[h]
\centering
\captionof{table}{Hyperparameter study on token length.}
\label{tab:ablation_token}
\begin{tabular}{ccc}
\toprule
\textbf{Length of token} & \textbf{PSNR↑} & \textbf{SSIM↑} \\
\midrule
10 & 35.2440  & 0.9762  \\
25 & 35.2552 & \textbf{0.9776}  \\
50 & \textbf{35.2558} & 0.9774 \\
\bottomrule
\end{tabular}
\end{table}
The group number $G$ controls the granularity of the PhySe-RPO.
As shown in Table~\ref{tab:ablation_G}, small $G$ values reduce exploration diversity and may lead to unstable reward estimation, limiting the benefits of PhySe-RPO. 
As $G$ increases, the model receives more reliable group-level comparisons, enabling more stable and effective policy updates. 
However, an excessively large $G$ value will introduce additional computational load and memory overhead. 
Our experiments show that moderate group sizes strike the best balance between exploration, stability, and computational cost, providing the strongest performance.
\begin{table}[t!]
\centering
\caption{Hyperparameter study on number of groups $G$.}
\label{tab:ablation_G}
\setlength{\tabcolsep}{2.5pt} 
\resizebox{\columnwidth}{!}{
\begin{tabular}{cccccccc}
\toprule
\footnotesize \textbf{Number of G} & \footnotesize \textbf{SSEQ↓} & \footnotesize \textbf{MANIQA↑} & \footnotesize \textbf{PI↓} & \footnotesize \textbf{FADE↓} & \footnotesize \textbf{MUSIQ↑} & \footnotesize \textbf{IS↑} & \footnotesize \textbf{NIQE↓}\\
\midrule
 2 & 4.036 & 0.346 & 3.296 & 0.279 & 53.751 & 2.796 & 4.777\\
\textbf{4} & \textbf{3.443} & \textbf{0.378} & \textbf{3.125} & \textbf{0.216}& 54.911 & 2.797 & \textbf{4.608} \\
 8 & 3.517 & 0.374 & 3.235 & 0.232 & \textbf{55.125} & \textbf{2.806} & 4.657 \\
\bottomrule
\end{tabular}
}
\end{table}


\end{document}